\documentclass[pdflatex,sn-mathphys-num]{sn-jnl}% Math and Physical Sciences Numbered Reference Style
\usepackage[monochrome]{xcolor}

\usepackage{graphicx}%
\usepackage{multirow}%
\usepackage{amsmath,amssymb,amsfonts}%
\usepackage{amsthm}%
\usepackage{mathrsfs}%
\usepackage[title]{appendix}%
\usepackage{xcolor}%
\usepackage{textcomp}%
\usepackage{manyfoot}%
\usepackage{booktabs}%

\usepackage{listings}%

\usepackage{algorithm}
\usepackage{algorithmic}

\usepackage{bbding}
\usepackage{float}
\usepackage{subcaption}
\usepackage{cleveref}

\theoremstyle{thmstyleone}%
\theoremstyle{thmstyletwo}%

\theoremstyle{thmstylethree}%

\begin{document}

\title[Article Title]{DCSCR: A Class-Specific Collaborative Representation based Network for Few-Shot Image Set Classification}


\author*[1]{\fnm{Xizhan} \sur{Gao}}\email{ise\_gaoxz@ujn.edu.cn}
\equalcont{These authors contributed equally to this work.}

\author[1]{\fnm{Wei} \sur{Hu}}
\equalcont{These authors contributed equally to this work.}

\affil[1]{\orgdiv{School of Information Science and Engineering}, \orgname{University of Jinan}, \orgaddress{\city{Jinan}, \postcode{250022}, \state{Shandong}, \country{China}}}


\abstract{
Image set classification (ISC), which can be viewed as a task of comparing similarities between sets consisting of unordered heterogeneous images with variable quantities and qualities, has attracted growing research attention in recent years. How to learn effective feature representations and how to explore the similarities between different image sets are two key yet challenging issues in this field. However, existing traditional ISC methods classify image sets based on raw pixel features, ignoring the importance of feature learning. Existing deep ISC methods can learn deep features, but they fail to adaptively adjust the features when measuring set distances, resulting in limited performance in few-shot ISC. To address the above issues, this paper combines traditional ISC methods with deep models and proposes a novel few-shot ISC approach called Deep Class-specific Collaborative Representation (DCSCR) network to simultaneously learn the frame- and concept-level feature representations of each image set and the distance similarities between different sets. Specifically, DCSCR consists of a fully convolutional deep feature extractor module, a global feature learning module, and a class-specific collaborative representation-based metric learning module. The deep feature extractor and global feature learning modules are used to learn (local and global) frame-level feature representations, while the class-specific collaborative representation-based metric learning module is exploit to adaptively learn the concept-level feature representation of each image set and thus obtain the distance similarities between different sets by developing a new CSCR-based contrastive loss function. Extensive experiments on several well-known few-shot ISC datasets demonstrate the effectiveness of the proposed method compared with some state-of-the-art image set classification algorithms.
}




\maketitle

\section{Introduction}
\label{sec:intro}

Image set classification (ISC) \cite{Guan2024RGMMC, Zhao2024VMHI} is an important research direction in computer vision filed, and in recent years, it has attracted much attention from researchers. Unlike traditional single image based classification methods, ISC methods aim to compare similarly between image sets with variable quantity, quality and unordered heterogeneous images. Although more information can be used within a set, ISC remains challenging due to the large intra-class variations as well as small inter-class differences in samples within the set captured in unconstrained environments. Hence, how to learn effective feature representations and how to explore the similarities between different image sets are two key yet challenging problems in ISC task. With the effort of researchers, many ISC methods have been proposed, and these methods can be grouped into two categories: traditional ISC methods and deep ISC methods.

Traditional ISC methods usually focus on learning the concept-level feature representations, i.e., focus on studying the modeling approaches of image set. According to the modeling approaches, existing traditional ISC methods can be further divided into parametric methods and non-parametric methods. Parametric methods (such as Manifold Density Divergence \cite{arandjelovic2005face}, etc.) model image sets using statistical distributions, while non-parametric methods model image sets using linear or non-linear subspaces. In recent years, the theory of representation learning based on the reconstruction principle has been greatly developed, mainly including sparse representation and collaborative representation. With the evolving demands of data processing, these methods have been applied to ISC task in order to learn more effective distance metrics, and they have shown surprising results in few-shot ISC. The representative representation learning based ISC methods include Image Set Collaborative Representation Classification algorithm (ISCRC) \cite{zhu2014image}, Dual Linear Regression Classification (DLRC) \cite{chen2014dual}, Pairwise Linear Regression Classification (PLRC) \cite{feng2016pairwise}, Joint Metric Learning-based Class-specific representation (JMLC) \cite{gao2022joint}, etc. These methods use affine or convex hulls to model image sets, use collaborative representation or class-specific collaborative representation (CSCR) theory to adaptively compute the weights of all images, and have achieved satisfactory classification performance. However, the aforementioned traditional ISC methods typically use the raw pixel values of images as their features, which imposes strict requirements on image quality and often exhibits limited discriminative ability. In other words, this type of method ignores the learning of frame-level features.

\begin{figure}[t]
\centering
\includegraphics[width=0.9\columnwidth]{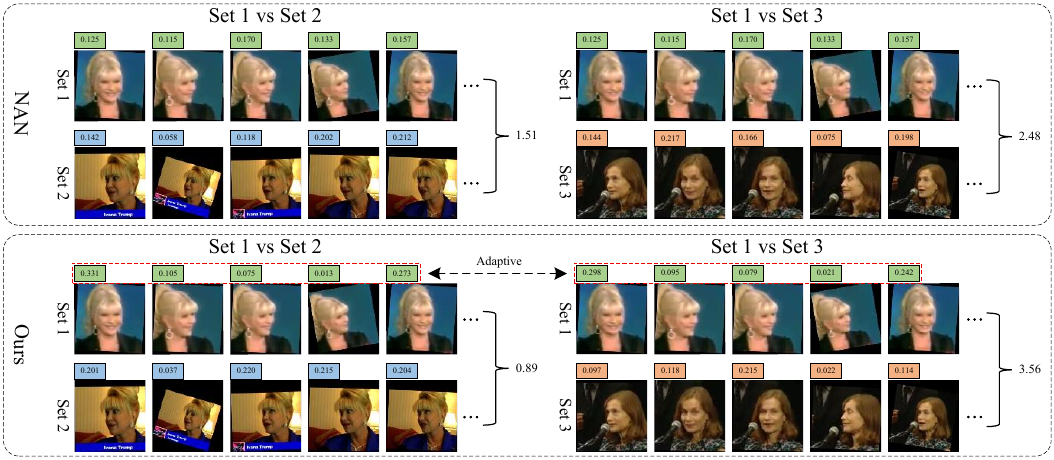}
\caption{Examples of weights of different methods on YTF dataset. NAN independently outputs fixed aggregation weight coefficients for each image set, while our method adaptively adjusts these weight coefficients when calculating distance. Besides, our method outputs a larger distance gap, which enables it to match the image set more accurately.}
\label{fig:motivation}
\end{figure}

Recently, deep ISC methods such as GhostVLAD \cite{zhong2019ghostvlad} and Neural Aggregation Network (NAN) \cite{yang2017neural} have appeared. These methods integrate Convolutional Neural Networks (CNNs) with ISC tasks, and have demonstrated the potential of deep image set-based classification algorithms. However, part of deep ISC algorithms focus on learning the frame-level feature representations, ignoring the learning of concept-level features. Another part algorithms aggregate the image set into a fixed feature vector (i.e., concept-level feature representation), which inevitably limits their representational capabilities (as shown in Fig. \ref{fig:motivation}). Because the ideal concept-level feature representation of an image set should be automatically adjusted based on the image set to be compared. Besides, convolutional layers used in these deep ISC methods can only capture the local features, ignoring the importance of global feature learning. Furthermore, experimental results show that these models perform well on large-scale datasets, but they exhibit obvious overfitting and poor generalization ability on few-shot image set classification datasets.

To simultaneously learn the frame- and concept-level feature representations of each image set and the distance similarities between different sets, this paper proposes a novel few-shot ISC approach called Deep Class-Specific Collaborative Representation (DCSCR) network, which combines deep neural networks with class-specific collaborative representation image set classification methods, thus integrating the concept-level modeling advantages of traditional methods with the powerful frame-level feature learning abilities of deep networks. In other words, this approach aims to fully utilize the rich information inherent within image sets, resulting in more discriminative deep adaptive concept-level representations. Specifically, the method first uses a deep convolutional neural network to learn a local frame-level feature representation of each image within the image set. Then, a global feature learning module is developed to adaptively learn a global frame-level feature representation for each image. Finally, a class-specific collaborative representation-based metric learning module is designed, which contains virtual modeling layers and CSCR-based contrastive loss function, to adaptively learn the concept-level feature representation of an image set. The main contributions of this article are as follows:

\hangafter 1 \hangindent 1.45em \noindent % 悬挂缩进
$\bullet$ A novel DCSCR method is proposed for ISC. In contrast to the conventional algorithms, this new method makes full use of the concept-level modeling advantages of traditional methods and the powerful frame-level feature learning capabilities of deep networks.
    
\hangafter 1 \hangindent 1.45em \noindent % 悬挂缩进
$\bullet$ A global feature learning module is used to summarize information from all pixels in an image, which can adaptively learn a global frame-level feature representation for each image. 
    
\hangafter 1 \hangindent 1.45em \noindent % 悬挂缩进
$\bullet$ A class-specific collaborative representation-based metric learning module is proposed. In contrast to existing deep ISC methods, this module can automatically adjust the concept-level features when computing image set distances, thus obtaining a more accurate distance metric.

The rest of the paper is organized as follows. In Section \ref{sec:related}, we briefly review the existing image set classification approaches. In Section \ref{sec:method}, we present the DCSCR framework and propose the bi-level training process. The experimental results of our method are shown in Section \ref{sec:experiments}. Section \ref{sec:conclusion} concludes the paper.

%%%%%%%%%%%%%%%%%%%%%%%%%%%%%%%%%%%%%%%%%%%%%%%%%%%%%%%%%%%%%%%%%%%%%%%%%%%%%%%%
\section{Related Work} \label{sec:related}

In this section, we briefly review some existing work about ISC methods. As mentioned above, existing image set classification methods can be roughly grouped into two types: traditional ISC methods and deep ISC methods. 

\subsection{Traditional Methods}
Classical traditional ISC methods \cite{yan2019multiple} usually use statistical features or subspaces to represent or model image sets, considering these models as points on manifolds. Specifically, Fukui et al. proposed an ISC framework based on a convex cone model that accurately represents the geometric structure of image sets \cite{sogi2022constrained}. Grassmann Discriminant Analysis \cite{hamm2008grassmann}, Grassmann Graph Embedding Discriminant Analysis \cite{harandi2011graph}, and other methods treated subspaces as points on the Grassmann manifold, and they learned discriminant functions based on kernel methods on this manifold. However, the drawback of these methods is the difficulty of obtaining explicit low-dimensional manifold structures. In addition, to obtain optimal tangent projections, Huang et al. proposed the Log-Euclidean Metric Learning (LEML) method \cite{huang2015log}. This method mapped the image sets from the original tangent space to a more discriminative tangent space, thereby improving the classification performance. Harandi et al. \cite{harandi2017dimensionality} achieved more discriminative low-dimensional SPD manifolds  by learning orthogonal projections on the SPD manifold. Projection Metric Learning (PML) \cite{huang2015projection} learned low-dimensional embeddings directly on the Grassmann manifold, making the projected low-dimensional manifold more discriminative and useful for classification. 

\textcolor{red}{In recent years, the theory of representation learning based on the reconstruction principle has been greatly developed, mainly including sparse representation and collaborative representation. With the evolving demands of data processing, these methods have been applied to ISC task in order to learn more effective distance metrics, and they have shown surprising results in few-shot ISC. The representative representation learning based ISC methods include Image Set Collaborative Representation Classification algorithm (ISCRC) \cite{zhu2014image}, Dual Linear Regression Classification (DLRC) \cite{chen2014dual}, Pairwise Linear Regression Classification (PLRC) \cite{feng2016pairwise}, Joint Metric Learning-based Class-specific representation (JMLC) \cite{gao2022joint}, etc. These methods use affine or convex hulls to model image sets, use collaborative representation or class-specific collaborative representation (CSCR) theory to adaptively compute the weights of all images, and have achieved satisfactory classification performance.} 

\textcolor{red}{Despite the ability to achieve adaptive image set modeling, traditional collaborative representation based ISC methods rely on low-level handcrafted features (e.g., raw pixel values) and lack the ability to learn discriminative frame-level visual features, which limits their generalization in complex unconstrained scenarios. In other words, these methods excel at concept-level feature learning yet neglect the importance of frame-level feature learning within image sets. This inherent limitation of traditional collaborative representation methods inspires our research motivation: to integrate the adaptive concept-level modeling advantage of collaborative representation with the powerful frame-level feature learning capability of deep neural networks, thus making up for the deficiencies of traditional methods.}

\subsection{Deep Methods}
Currently, research on deep learning for ISC \cite{wang2022learning, wang2023u} is in its early stages. The ISC algorithm GhostVLAD \cite{zhong2019ghostvlad}, proposed by DeepMind and VGG, used deep networks to aggregate multiple face images within an image set into a compact and discriminative feature vector. The GaitSet algorithm \cite{chao2019gaitset} treated gait sequences as unordered sets of images and introduced a set-pooling strategy to aggregate features from multiple images extracted by CNN networks into a single feature vector for gait recognition. The Neural Aggregation Network (NAN) \cite{yang2017neural} was inspired by attention mechanisms and adaptively learns image set modeling as a fixed-length feature vector. In addition, to address the problem of inconsistent image quality in set-to-set recognition tasks, Liu et al. proposed a Quality-Aware Network (QAN) \cite{liu2017quality} that automatically learns the quality scores of each sample, allowing weighted aggregation of features based on different sample qualities. 

\textcolor{red}{Despite the progress made by these deep ISC methods, they suffer from a core limitation: the lack of adaptive concept-level modeling due to their fixed feature aggregation/set-pooling strategies (which aggregate an image set into a fixed feature vector). These fixed mechanisms fail to adjust concept-level features according to paired image sets, which limits their representational capability and generalization performance in few-shot ISC tasks. Recent studies have shown that hybrid deep learning architectures combined with metaheuristic-based hyperparameter optimization achieve excellent performance in complex pattern recognition tasks \cite{shukla2025fraudulent}. Inspired by this, we propose the DCSCR network, which incorporates the CSCRMLM module to realize adaptive joint modeling of paired image sets. This module dynamically adjusts the concept-level feature weights based on the mutual information of paired sets, thus making up for the deficiencies of existing deep ISC models and improving the performance on the few-shot ISC tasks.}

\begin{figure}[t]
\centering
\includegraphics[width=0.9\columnwidth]{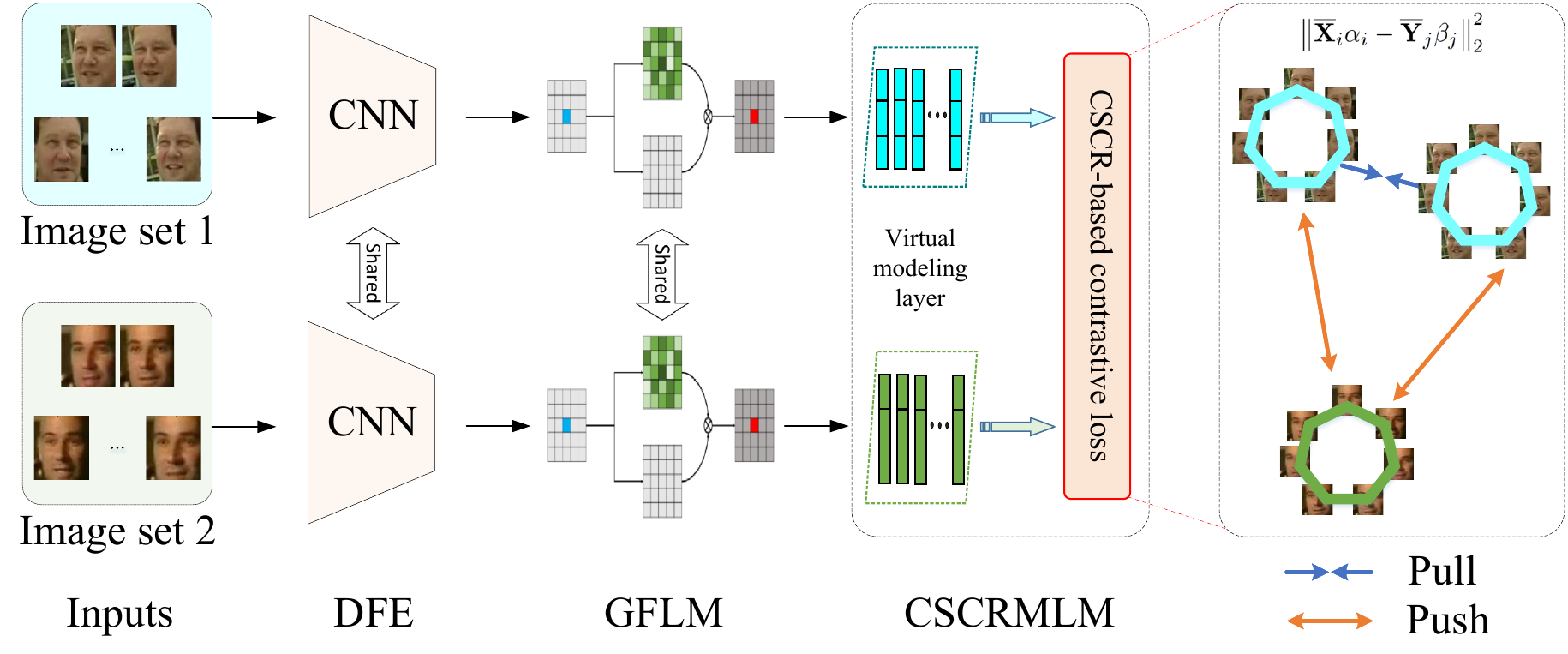}
\caption{The diagram of the proposed DCSCR network, which consists of a fully convolutional DFE, a GFLM, and a CSCRMLM. \textcolor{red}{For input image sets, we sequentially adopt DFE and GFLM to learn local frame-level features and global frame-level features, respectively. Then, we employ CSCRMLM to perform adaptive modeling on the image sets, and finally obtain the concept-level features and the distance between different sets.}}
\label{fig1}
\end{figure}

%%%%%%%%%%%%%%%%%%%%%%%%%%%%%%%%%%%%%%%%%%%%%%%%%%%%%%%%%%%%%%%%%%%%%%%%%%%%%%%%
\section{Proposed Method} \label{sec:method}

The image set classification problem can be viewed as matching and comparing two image sets. Formally, we define two training image sets as $\mathbf{G}=[g_{1}, g_{2}, \cdots, g_{m}]\in \mathbb{R}^{D \times m}$, and $\mathbf{P}=[p_{1}, p_{2}, \cdots, p_{n}]\in \mathbb{R}^{D \times n} $. In this case, $\mathit{D} $ is the dimension of the image features, \textcolor{red}{and $\mathit{m, n}$ represent the number of images in these two image sets,} respectively. 

As illustrated in Fig. \ref{fig1}, to simultaneously learn the frame- and concept-level feature representations of each image set and the distance similarities between different sets, our deep class-specific collaborative representation approach is designed to contain three parts: a fully convolutional deep feature extractor (DFE), a global feature learning module (GFLM), and a class-specific collaborative representation-based metric learning module (CSCRMLM). \textcolor{red}{The DFE module is used to learn the local frame-level features of each image. Then, the GFLM is utilized to capture global frame-level features. After that, we employ the CSCRMLM to adaptively learn concept-level feature representations and measure the distance similarities between different sets.}

\subsection{Deep Feature Extractor}
The DFE module is first designed to learn local frame-level features. More specifically, to better capture the discriminative information from the images, we select the convolutional layers of advanced deep neural networks (such as ResNet50 \cite{he2016deep}, GoogleNet \cite{szegedy2016rethinking}, etc.), which have demonstrated superior classification performance, as the feature extractor. By using convolutional networks, images are mapped into a discriminative space, and corresponding feature sets are generated for the above two image sets, denoted as $\mathbf{X} =\left \{X_{i}\right \} _{i=1}^{m} $ and $\mathbf{Y} =\left \{Y_{j}\right \} _{j=1}^{n}$, respectively. Here, $X_{i}$ and $Y_{j}$ are feature maps with dimensions $H \times W \times D$, where H, W, and D represent the height, width, and channel dimensions of the feature maps. After training the feature extractor with a large-scale facial dataset, we will freeze its parameters.

\subsection{Global Feature Learning Module}
The convolution operation processes one local neighborhood at a time, so it cannot model dependencies that extend beyond the small receptive field. To capture the global dependencies at all spatial locations in an image (i.e., global frame-level features), we apply the self-attention mechanism to our framework. The self-attention mechanism is able to aggregate the information from each pixel in an image, allowing the network to focus on task-relevant features so that it can adaptively learn more discriminative feature representations.

Inspired by the aforementioned analysis, we use the non-local block, proposed by Wang et al. \cite{wang2018non}, to construct our self-attention module. After the self-attention operation, we aggregate the features and generate the final feature representations for the two image sets $\mathbf{X}$ and $\mathbf{Y}$ by applying the Global Average Pooling (GAP) operation, where $x_{m}^{'}$ and $y_{n}^{'}$ are the outputs of self-attention module:
\begin{equation}
\begin{cases} 
 & \mathbf{\overline{X}} =\left \{ \overline{x}_{1}, \overline{x}_{2}, \cdots, \overline{x}_{m}  \right \}, \overline{x}_{m}  = \rm{GAP}(\it{x}_{m}^{'})\in \mathbb{R}^{\rm{1\times 1\times D}},
\\ & \mathbf{\overline{Y}} =\left \{ \overline{y}_{1}, \overline{y}_{2}, \cdots, \overline{y}_{n}  \right \}, \overline{y}_{n}  = \rm{GAP}(\it{y}_{n}^{'})\in \mathbb{R}^{\rm{1\times 1\times D}}.
\end{cases}
\end{equation}

The formula of the non-local block shows that it supports arbitrary length inputs and keeps the length and feature dimension of the input during output. Therefore, the non-local block can be flexibly added to any pre-trained feature extractor and fine-tuned together with it.

\subsection{CSCR-based Metric Learning Module}
Through the above two modules, only frame-level feature representations can be obtained, our next goal is to learn concept-level feature representations. As aforementioned, we think the ideal concept-level feature representation of an image set should be automatically adjusted based on the image set to be compared. Hence, \textcolor{red}{inspired by the adaptive modeling idea of traditional collaborative representation for ISC, we propose the CSCR-based metric learning module (CSCRMLM), in which} the class-specific collaborative representation (CSCR\footnote{Our previous works \cite{gao2022joint} have demonstrated the effectiveness of CSCR in image set classification field.}) is used to adaptively learn the concept-level feature representations and thus obtain the distance similarities between different sets. 

\textcolor{red}{CSCRMLM inherits the core advantage of collaborative representation in adaptively learning the weight coefficients of image samples for concept-level feature construction, and makes two key improvements to address the limitations of traditional methods. Different from traditional collaborative representation methods that use low-level handcrafted features, our CSCRMLM takes the deep frame-level features (local and global) learned by DFE and GFLM as input, which provides more discriminative visual information for concept-level modeling. In addition, we design a CSCR-based contrastive loss function to integrate the adaptive weight learning of CSCR into the end-to-end deep network training, realizing the joint optimization of frame-level feature learning and concept-level metric learning, which is very different from the separate feature extraction and metric learning of traditional collaborative representation methods. Finally, the CSCR-based metric learning module is designed to consist of two components: a virtual modeling layer and a CSCR-based contrastive loss function.}

\textbf{Virtual modeling layer.} Assuming that there is a virtual image $\mathit{v}$ at the intersection of two spanning subspaces $\mathbf{\overline{X}}$, $\mathbf{\overline{Y}}$, so it can be modelled by these subspaces as $v_{x}=\mathbf{\overline{X}}{\alpha}$ and $v_{y}=\mathbf{\overline{Y}}{\beta}$, based on the definition of spanning subspaces. Thus, the distance between the two image sets can be replaced by the distance between $v_{x}$ and $v_{y}$. Here, the concept-level feature representations, i.e. $\mathbf{\overline{X}}{\alpha}$ and $\mathbf{\overline{Y}}{\beta}$, can be seen as the outputs of a virtual modeling layer.

\textbf{CSCR-based contrastive loss.} After obtaining the concept-level feature representations, theoretically we can compute the distance between two image sets. However, at this point, $\alpha$ and $\beta$ are unknown parameters that need to be optimized and solved. To solve the above problem and guide the network training, we proposed the following CSCR-based contrastive loss:
\begin{equation}
\label{loss}
\begin{aligned}
\mathit{L} = \ & y_{ij}\mu_{1}\left \Vert \mathbf{\overline{X}}_{i}\alpha_{i}-\mathbf{\overline{Y}}_{j}\beta_{j} \right \Vert _{2}^{2}
+(1-y_{ij})\mu_{2}\rm{max}(0, \textcolor{red}{\it{t} }-\left \Vert \mathbf{\overline{X}}_{\it{i}}\alpha_{\it{i}}-\mathbf{\overline{Y}}_{\it{j}}\beta_{\it{j}} \right \Vert _{\rm{2}}^{\rm{2}})
\\&+\lambda_{1} \left \Vert \alpha_{i} \right \Vert_{2}^{2} +\lambda_{2}\left \Vert \beta_{j} \right \Vert _{2}^{2}
\\&s.t. \sum_{k}\alpha_{i,k}=1, \sum_{k}\beta_{j,k}=1.
\end{aligned}
\end{equation}
Here, $\mathbf{\overline{X}}_{i}$ and $\mathbf{\overline{Y}}_{j}$ are two image sets, and $y_{ij}$ is the indicates function, where $y_{ij}=1$ indicates that the two sets belong to the same category, and $y_{ij}=0$ indicates that the two sets belong to different categories. \textcolor{red}{$\mu_1$ and $\mu_2$ denote the weight coefficients of the contrastive loss for intra-class and inter-class image set pairs, respectively, $\lambda_1$ and $\lambda_2$ are the $l_2$ regularization coefficients for the collaborative representation weights $\alpha_i$ and $\beta_j$. The constant $\mathit{t}$ is the threshold,} and in our experiments it is set to 2.

The first two $l_2$ norm regularization terms allow the affine subspaces to emphasize high quality images and suppress the contribution of low quality images to the affine subspace modeling. \textcolor{red}{For example, if $\mathbf{\overline{X}}_{i}$ and $\mathbf{\overline{Y}}_{j}$ belong to the same category, the second regularization term becomes $0$, and only the first regularization term takes effect. Although the image sets $\mathbf{\overline{X}}_{i}$ and $\mathbf{\overline{Y}}_{j}$ usually contain both high quality and low quality images (see Fig. \ref{fig:motivation}), high quality images within the same category are often highly similar, while large discrepancies exist between low quality images of the same category and between low quality and high quality images. Driven by the optimization objective of the first regularization term to minimize $\left \Vert \mathbf{\overline{X}}_{i}\alpha_{i}-\mathbf{\overline{Y}}_{j}\beta_{j} \right \Vert _{2}^{2}$, the model automatically prioritizes high quality images from $\mathbf{\overline{X}}_{i}$ and $\mathbf{\overline{Y}}_{j}$ for modeling, thereby naturally achieving the effect of emphasizing high quality images and suppressing low quality images during feature learning. This mechanism also holds when $\mathbf{\overline{X}}_{i}$ and $\mathbf{\overline{Y}}_{j}$ belong to different categories.}

The third and fourth regularization terms are used to reconstruct the affine subspace with as few samples as possible. \textcolor{red}{In other words, by minimizing the $l_2$ norm of the coefficients \((\alpha_i, \beta_j)\), the model encourages reconstruction using fewer and higher quality (more discriminative) samples, thereby reducing low quality frames.} The constraint $\sum_{k}\alpha_{i,k}=1$ and $\sum_{k}\beta_{j,k}=1$ is used to avoid the trivial solution. Finally, the distance between the two sets can be computed using $\left \Vert \mathbf{\overline{X}}_{i}\alpha_{i}-\mathbf{\overline{Y}}_{j}\beta_{j} \right \Vert_{2}^{2}$. This loss function explicitly encourages two sets from the same category to be close to each other, and two sets from different categories to be far from each other.

The main difference between our proposed method and existing works is that: existing works usually construct aggregation networks (such as attention mechanisms or LSTM) to learn aggregation coefficients or concept-level modeling coefficients. However, they only consider the images in one set during modeling, ignoring the information exchange between different sets. In theory, when calculating the distance between two sets, their feature representations are interdependent. In addition, building an aggregation network will increase the number of network parameters and affect computational efficiency. On the contrary, our proposed method can adaptively calculate the modeling coefficients of two image sets without introducing additional parameters, resulting in faster training/inference speed and accurate distance measurement. To jointly train the proposed DCSCR method, we designed the following bi-level training and alternative optimization scheme.

\subsection{Bi-level Training and Alternative Optimization} \label{sec:opti}
\textcolor{red}{To effectively train the entire DCSCR network, we adopt a bi-level training strategy. As shown in \textbf{Algorithm \ref{alg:DCSCR}}, we use the pre-trained ResNet50 as our backbone network. In the Level-1 training stage, the GFLM is initially trained using the cross-entropy loss. In the Level-2 training stage, the proposed CSCR-based contrastive loss is employed to alternately optimize the GFLM and the CSCRMLM.}

\subsubsection{\textcolor{red}{Level-1 Training}}
In the first training stage, we focus on pre-train the deep feature extractor and global feature learning modules. Specifically, we perform global average pooling (GAP) operation on the features obtained from the global feature learning module, resulting in a fixed-size representation (2048-d). Then, we add a Softmax layer behind GAP to train the model using cross-entropy loss. 

\subsubsection{\textcolor{red}{Level-2 Training}}
In the second training stage, we add the CSCR-based metric learning module and use the proposed contrastive loss function in Eq. (2) to guide the model training. More specifically, we use the following optimization scheme to incorporate the parameter-free collaborative representation algorithm into the deep model:

$\textbf{Step 1}$: Fix the neural network parameters $\theta$ and solve the optimization problem with respect to $\alpha$ and $\beta$. The task in this step is to find the parameters $\alpha$ and $\beta$ for collaborative representation. Depending on the different values of $y_{ij}$, our loss function can be divided into two sub-problems:

\begin{equation}
\label{loss1}
\begin{aligned}
& \min_{\alpha_{i},\beta_{j}} \  \mu_{1}\left \Vert \mathbf{\overline{X}}_{i}\alpha_{i}-\mathbf{\overline{Y}}_{j}\beta_{j} \right \Vert _{2}^{2}+\lambda_{1}\left \Vert \alpha_{i} \right \Vert_{2}^{2} +\lambda_{2}\left \Vert \beta_{j} \right \Vert _{2}^{2}   \\
& s.t. \  \sum_{k}\alpha_{i,k}=1, \  \sum_{k}\beta_{j,k}=1,  \  y_{ij}=1.
\end{aligned}
\end{equation}

\begin{equation}
\label{loss2}
\begin{aligned}
& \min_{\alpha_{i},\beta_{j}}  \  \mu_{2}\left \Vert \mathbf{\overline{X}}_{\it{i}}\alpha_{\it{i}}-\mathbf{\overline{Y}}_{\it{j}}\beta_{\it{j}} \right \Vert _{\rm{2}}^{\rm{2}}+\lambda_{\rm{1}}\left \Vert \alpha_{\it{i}} \right \Vert _{\rm{2}}^{\rm{2}} +\lambda_{\rm{2}}\left \Vert \beta_{\it{j}} \right \Vert _{\rm{2}}^{\rm{2}}   \\
& s.t. \  \sum_{k}\alpha_{i,k}=1, \  \sum_{k}\beta_{j,k}=1, \  y_{ij}=0.
\end{aligned}
\end{equation}
These two problems can be solved by using an iterative optimization approach based on the Alternating Direction Method of Multipliers (ADMM) \cite{wen2010alternating}. Besides, since these two problems are almost the same, we only take problem (\ref{loss1}) as an example. \textcolor{red}{To solve this problem, we first construct the following augmented Lagrangian function:}
\begin{equation}
\label{eq:3-4-3}
\begin{aligned}
L(\alpha_{i},\beta_{j}, \lambda_{1}, \lambda_{2}) = &\mu_{1}\left \Vert \mathbf{\overline{X}}_{i}\alpha_{i}-\mathbf{\overline{Y}}_{j}\beta_{j} \right \Vert _{2}^{2}+\lambda_{1}\left \Vert \alpha_{i} \right \Vert_{2}^{2} +\lambda_{2}\left \Vert \beta_{j} \right \Vert _{2}^{2} \\+ 
&\frac{\rho}{2}\left \Vert e_{\alpha}^{i^{T}}\alpha_{i}-1+\frac{\eta_{1}}{\rho} \right\Vert_{2}^{2} + 
\frac{\rho}{2}\left \Vert e_{\beta}^{j^{T}}\beta_{j}-1+\frac{\eta_{2}}{\rho} \right\Vert_{2}^{2}, 
\end{aligned}
\end{equation}
where $e_{\alpha}^{i}$ and $e_{\beta}^{j}$ are column vectors with all values 1 (i.e. $e_{\alpha}^{i}=[1, \cdots, 1]^{T}$,  $e_{\beta}^{j}=[1, \cdots, 1]^{T}$), the initial value of $\eta_{1}$ and $\eta_{2}$ is set to 0, $\lambda_{1}$ and $\lambda_{2}$ are two Lagrange multipliers, and $\rho$ is the coefficient of the regular term. \textcolor{red}{This is a non-convex optimization problem and is difficult to solve in closed form. Therefore, we adopt an alternating optimization strategy, i.e., fixing $\beta_j$ to solve $\alpha_i$ and fixing $\alpha_i$ to solve $\beta_j$. }

\textcolor{red}{When $\beta_j$ is fixed, the optimization problem (\ref{eq:3-4-3}) reduces to the following function:}
\begin{equation}
\label{eq:3-4-4}
\textcolor{red}{ 
\begin{aligned}
L_1 = &\mu_{1}\left \Vert \mathbf{\overline{X}}_{i}\alpha_{i}-\mathbf{\overline{Y}}_{j}\beta_{j} \right \Vert _{2}^{2}+\lambda_{1}\left \Vert \alpha_{i} \right \Vert_{2}^{2} + 
\frac{\rho}{2}\left \Vert e_{\alpha}^{i^{T}}\alpha_{i}-1+\frac{\eta_{1}}{\rho} \right\Vert_{2}^{2}.
\end{aligned} }
\end{equation}
\textcolor{red}{ By setting $\frac{\partial L_1}{\partial \alpha_i} =0$, the optimal value of $\alpha_i$ can be computed as:  }
\begin{equation}
\label{eq:3-4-5}
\alpha_{i}^{(k)} = P_{x}(2\mu_{1}\overline{X}_{i}^{T}\overline{Y}_{j}\beta_{j}^{(k-1)}+\rho e_{\alpha}^{i}-\eta_{1}^{(k-1)}e_{\alpha}^{i}),
\end{equation}
where $P_{x}=(2\mu_{1}\overline{X}_{i}^{T}\overline{X}_{i}+\rho e_{\alpha}^{i} e_{\alpha}^{i^{T}}+2\lambda_{1}I)^{-1}$, $I$ is an identity matrix, and $k$ is the number of iterations.

\textcolor{red}{When $\alpha_i$ is obtained, the optimization problem (\ref{eq:3-4-3}) reduces to the following function:}
\begin{equation}
\label{eq:3-4-6}
\textcolor{red}{ 
\begin{aligned}
L_2 = &\mu_{1}\left \Vert \mathbf{\overline{X}}_{i}\alpha_{i}-\mathbf{\overline{Y}}_{j}\beta_{j} \right \Vert _{2}^{2} + \lambda_{2}\left \Vert \beta_{j} \right \Vert _{2}^{2} +  
\frac{\rho}{2}\left \Vert e_{\beta}^{j^{T}}\beta_{j}-1+\frac{\eta_{2}}{\rho} \right\Vert_{2}^{2}.
\end{aligned}  }
\end{equation}
\textcolor{red}{ By setting $\frac{\partial L_2}{\partial \beta_j} =0$, the optimal value of $\beta_j$ can be computed as:  }
\begin{equation}
\label{eq:3-4-7}
\beta_{j}^{(k)} = P_{y}(2\mu_{1}\overline{Y}_{j}^{T}\overline{X}_{i}\alpha_{i}^{(k)}+\rho e_{\beta}^{j}-\eta_{2}^{(k-1)}e_{\beta}^{j}),
\end{equation}
where $P_{y}=(2\mu_{1}\overline{Y}_{j}^{T}\overline{Y}_{j}+\rho e_{\beta}^{j} e_{\beta}^{j^{T}}+2\lambda_{2}I)^{-1}$, $I$ is an identity matrix.

Finally, we can update the $\eta_{1}$ and $\eta_{2}$ as follows:

\begin{equation}
\label{eq:3-4-8}
\begin{aligned}
\eta_{1}^{(k)} = \eta_{1}^{(k-1)} + \rho(e_{\alpha}^{i^{T}}\alpha_{i}^{(k)}-1) \\
\eta_{2}^{(k)} = \eta_{2}^{(k-1)} + \rho(e_{\beta}^{j^{T}}\beta_{j}^{(k)}-1)
\end{aligned}
\end{equation}

Repeat the process (\ref{eq:3-4-5})-(\ref{eq:3-4-8}) until the end condition of the algorithm is satisfied, we can get the best $\alpha_{i}$ and $\beta_{j}$. \textcolor{red}{The ADMM optimization procedure is summarized in \textbf{Algorithm \ref{alg:ADMM}}. } In the similar manner, we can solve the problem (\ref{loss2}).

\begin{algorithm}[!t]
	\renewcommand{\algorithmicrequire}{\textbf{Input:}}
	\renewcommand{\algorithmicensure}{\textbf{Output:}}
    \renewcommand{\algorithmicwhile}{\textbf{While}}
    \renewcommand{\algorithmicendwhile}{\textbf{End while}} % 保持 end while 样式
	\caption{ADMM algorithm}
	\label{alg:ADMM}
\textcolor{red}{
	\begin{algorithmic}[1]
		\REQUIRE Training sets and labels; $\mu_{1}$; $\lambda_{1}$; $\lambda_{2}$; $\rho$
		\ENSURE The computed $\alpha_i$ and $\beta_j$
		\STATE Initialize $\beta_j = [1, ..., 1]^T$;
        \WHILE{not converged}
            \STATE Update $\alpha_i$ by solving Eq. (\ref{eq:3-4-5});
            \STATE Update $\beta_j$ by solving Eq. (\ref{eq:3-4-7});
            \STATE Update $\eta_1$, $\eta_2$ by solving Eq. (\ref{eq:3-4-8});
        \ENDWHILE
        \STATE Return $\alpha_i$ and $\beta_j$ obtained from the final iteration.
	\end{algorithmic} }
\end{algorithm}

$\textbf{Step 2}$: \textcolor{red}{After solving for $\alpha$ and $\beta$, the loss function (\ref{loss}) can be computed as a definite value, from which the derivative of this loss function with respect to the network parameter $\theta$ can be further derived:}

\begin{equation}
\label{eq:3-4-9}
\textcolor{red}{
\frac{\partial L}{\partial \theta} = \frac{\partial \mathbf{\overline{X}}_{i}}{\partial \theta}  \frac{\partial L}{\partial \mathbf{\overline{X}}_{i}} + \frac{\partial \mathbf{\overline{Y}}_{j}}{\partial \theta}  \frac{\partial L}{\partial \mathbf{\overline{Y}}_{j}}.
}
\end{equation} 
\textcolor{red}{Since closed-form solutions exist for both $\frac{\partial L}{\partial \mathbf{\overline{X}}_{i}}$ and $\frac{\partial L}{\partial \mathbf{\overline{Y}}_{j}}$, the network parameter $\theta$ can be updated using the stochastic gradient descent (SGD) algorithm:}
\begin{equation}
\label{eq:3-4-10}
\textcolor{red}{
\theta = \theta - \lambda  \frac{\partial L}{\partial \theta}.
}
\end{equation} 

\textcolor{red}{The overall bi-level training is outlined in \textbf{Algorithm \ref{alg:DCSCR}}.} The benefits of utilizing both of the global feature learning module and class-specific collaborative representation-based metric learning module are demonstrated in experiments.

\begin{algorithm}[!t]
	\renewcommand{\algorithmicrequire}{\textbf{Input:}}
	\renewcommand{\algorithmicensure}{\textbf{Output:}}
	\caption{Bi-level training of DCSCR}
	\label{alg:DCSCR}
	\begin{algorithmic}[1]
		\REQUIRE Training sets and labels; $\mu_{1}$; $\mu_{2}$; $\lambda_{1}$; $\lambda_{2}$; $m$; Pre-trained DFE (ResNet50)
		\ENSURE Updated neural network parameters $\theta$
		\STATE $\textbf{Level 1: Pre-train GFLM}$
		\FORALL{DFE extracted training sets \textbf{X} and \textbf{Y}}
		\STATE Reconstruct  $\left \{X_{i}\right \} _{i=1}^{m}$, $\left \{Y_{j}\right \} _{j=1}^{n}$ to $\mathbf{X^{'}}$, $\mathbf{Y^{'}}$ using global feature learning module;
		\STATE Compute $\overline{\mathbf{X}}$ = GAP($\mathbf{X^{'}}$); $\overline{\mathbf{Y}}$  = GAP($\mathbf{Y^{'}}$);
		\STATE Compute cross entropy loss and update global feature learning module by SGD;
		\ENDFOR
		\STATE $\textbf{Level 2: Alternative optimization with GFLM and CSCRMLM}$
		\FORALL{$\overline{\mathbf{X}}$ and $\overline{\mathbf{Y}}$ pairs}
		\STATE Compute indicates function $y_{ij}$ with training set labels;
		\STATE Compute $\alpha$, $\beta$ using ADMM method;
		\STATE Compute the contrastive loss in Eq. (2);
		\STATE Update the neural network parameters $\theta$ by SGD.
		\ENDFOR
	\end{algorithmic}
\end{algorithm}

%%%%%%%%%%%%%%%%%%%%%%%%%%%%%%%%%%%%%%%%%%%%%%%%%%%%%%%%%%%%%%%%%%%%%%%%%%%%%%%%
\section{Experiments} \label{sec:experiments}

In this section, a large number of experiments are performed on two few-shot image set classification datasets (i.e., Honda/UCSD and CMU MoBo datasets) and a large image set classification dataset (i.e., YouTube Faces dataset), to verify the effectiveness of the proposed DCSCR method. 

As mentioned earlier, we use a bi-level training strategy to train our network. The network is trained with standard backpropagation and a SGD solver. The batch size, learning rate, iteration and hyper-parameters are tuned for each dataset.

\subsection{Baselines and Comparison Methods}

To evaluate the classification performance of our proposed network, we compare with some state-of-the-art ISC methods, including traditional ISC methods: DCC \cite{kim2008canonical}, MMD \cite{wang2012manifold}, etc., and deep ISC methods: NAN, FaceNet, DLMC \cite{wu2020angular}, and BDML \cite{wei2022discrete}. All methods are empirically tuned based on the recommendations in the original references and the source code provided by the original authors to achieve the best classification accuracy.

Our baseline methods include nonlocal+GAP and nonlocal+DFA. The nonlocal+GAP is a simplified version of our proposed method, i.e., it replaces the CSCRMLM part by a linear classifier. The nonlocal+DFA is a realization of PIFR \cite{liu2019permutation} by ourselves, i.e., it use a nonparametric dictionary learning method \textcolor{red}{to calculate the concept-level similarity.}

\textcolor{red}{To ensure the rationality and representativeness of the selected baselines for the few-shot ISC task, we make the selection from two core considerations. First, recent state-of-the-art deep ISC methods (e.g., SMDML \cite{Wang2024SMDML}, multi kernel metric learning framework (MKMLF) \cite{Wang2022MRMD}, SymNet \cite{Wang2022SymNet}, SSMAE \cite{Wang2024DMLSM}) are primarily designed and evaluated on large-scale datasets, with no publicly available experimental results or official codes under the few-shot setting, making direct and fair quantitative comparison infeasible. Second, the baselines and comparison methods we selected cover the two mainstream categories of ISC approaches (traditional and deep ISC methods), including classical representation learning methods (ISCRC, DLRC, PLRC), manifold-based methods (LEML, PML), and representative deep aggregation methods (NAN, FaceNet) that are widely adopted in few-shot ISC research. In addition, the self-constructed baselines (nonlocal+GAP, nonlocal+DFA) are designed to isolate the contribution of the core CSCRMLM module, which can directly verify the effectiveness of our proposed module by contrast. All selected methods are reproducible with publicly available codes or implementation guidelines, and we tune their hyper-parameters strictly following the original references to achieve optimal performance, ensuring the fairness and validity of the comparative experiments.}

\subsection{Datasets}
We conduct experiments on two few-shot image set classification datasets (i.e., Honda/UCSD and CMU MoBo datasets) and a large image set classification dataset (i.e., YouTube Faces dataset). 

$\textbf{The Honda/UCSD dataset}$ \cite{lee2003video} consists of 59 image sets from 20 different classes. Each set has approximately 200 to 600 frames with varying poses and expressions. To ensure a fair comparison, we randomly select one set from each class to build the gallery set and use the remaining as the probe sets. In other words, one image set per class is used for training, and this setup belongs to the standard few-shot classification problem. \textcolor{red}{The core variability of the samples in this dataset lies in the unconstrained variations in poses and expressions, with significant intra-class sample quality differences. Meanwhile, only one image set per class is used during the training phase, making it a typical few-shot image set classification dataset, which imposes high requirements on the feature robustness and generalization ability of the model.}

$\textbf{The CMU MoBo dataset}$ \cite{gross2001cmu} contains 87 image sets from 24 individuals walking on a treadmill. For each class, an image set is randomly selected to construct the training set, and the rest is used to construct the probe set. So, this is also a standard few-shot classification problem. \textcolor{red}{The sample variability of this dataset is mainly reflected in the subtle changes in posture and differences in shooting perspectives during treadmill walking, with strong inter-frame feature correlation. Since only one image set per class is used during the training phase, it is also a typical few-shot image set classification dataset.}

$\textbf{The YouTube Face (YTF) dataset}$ \cite{wolf2011face} is a large-scale ISC dataset, which has 3,425 videos from 1,595 different subjects. The dataset provides ten folds of 5000 video pairs, and each fold contains about 250 positive pairs and 250 negative pairs.  \textcolor{red}{The core sample variability of this dataset is reflected in the complex unconstrained variations in illumination, background and pose, with significant differences in the number of frames across different videos. This dataset can effectively verify the model's classification and verification capabilities in large-scale, high-variability scenarios.}

\subsection{Evaluation on Honda/UCSD Dataset}

For the Honda/UCSD dataset, we select first 50, 100, 200 frames from each set to conduct our experiments. The average classification accuracy of ten random experiments are reported in Table \ref{Honda/UCSD}. From the table, it can be seen that our proposed DCSCR method achieves the best results in most cases, which shows the effectiveness of our proposed framework. Specifically, on first 50 frames case, our method achieves the highest classification performance ($97.95\%$); on first 100 frames case and first 200 frames case, our method achieves the best classification performance ($100.0\%$). We notice that some methods are sensitive to the size of the image sets (such as CSRbDML, DLRC, and PLRC), with the increase of the number of images, their recognition performance decreases. In contrast, our proposed method does not have this shortcoming. Compared to the latest JMLC method, the proposed DCSCR achieves higher performance in all cases, demonstrating the effectiveness of DCSCR. 

In addition, compared with the baseline method nonlocal+GAP, our proposed method achieves much higher accuracy (about 20\% higher) in all cases, which can demonstrate the importance of CSCRMLM part. Compared with the baseline nonlocal+DFA (PIFR), we find again that our DCSCR achieves higher accuracy in all cases, which means that compared to traditional dictionary learning method, our class-specific collaborative representation-based metric learning module has more powerful learning ability.

Furthermore, compared with NAN and FaceNet methods that aggregate the image set into a fixed feature vector, the adaptive adjustment methods (i.e., DCSCR and PIFR) perform better in all cases. This phenomenon is consistent with the motivation of this paper: the concept-level features of an image set should be automatically adjusted based on the image sets to be compared.

\begin{table}[t]
\caption{Average classification accuracy and standard deviations on Honda dataset (\%).} \label{Honda/UCSD}
\centering
\scriptsize{  % \small, \footnotesize, \scriptsize, \tiny
\begin{tabular}{|c|c|c|c|}
\hline
\textbf{Methods} & \textbf{50 frames}& \textbf{100 frames}& \textbf{200 frames} \\
\hline
DCC \cite{kim2008canonical} & $86.67\pm3.16$ & $91.79\pm1.73$ & $95.38\pm6.48$ \\
MMD \cite{wang2012manifold} & $81.03\pm5.48$ & $86.41\pm6.93$ & $94.10\pm3.46$ \\
AHISD \cite{cevikalp2010face} & $86.25\pm2.45$ & $88.75\pm2.65$ & $91.00\pm3.02$ \\
CHISD \cite{cevikalp2010face} & $83.25\pm2.47$ & $88.00\pm3.32$  & $89.00\pm5.00$ \\
SANP \cite{hu2012face} & $81.07\pm8.37$ & $89.03\pm5.74$ & $92.71\pm3.32$ \\
RNP \cite{yang2013face} & $90.77\pm4.39$ & $94.62\pm3.30$ & $96.41\pm2.16$ \\
ISCRC \cite{zhu2014image} & $90.03\pm5.29$ & $92.01\pm1.74$ & $95.64\pm1.73$ \\
DLRC \cite{chen2014dual} & $88.21\pm3.61$ & $90.51\pm5.57$ & $84.10\pm4.58$ \\
LEML \cite{huang2015log} & $66.67\pm7.55$ & $92.31\pm3.16$ & $95.64\pm3.16$ \\
PML \cite{huang2015projection} & $92.33\pm3.16$ & $94.10\pm3.61$ & $96.15\pm1.84$ \\
PLRC \cite{feng2016pairwise} & $90.77\pm3.46$ & $92.36\pm7.81$ & $62.82\pm7.94$ \\
DRA \cite{ren2019discriminative} & $86.67\pm9.19$ & $89.24\pm4.65$ & $88.21\pm5.30$ \\
FaceNet \cite{schroff2015facenet} & $93.53\pm2.53$ & $98.94\pm0.82$ & $99.73\pm0.12$ \\
GCR \cite{liu2019group} & $92.82\pm3.34$ & $96.92\pm4.22$ & $98.46\pm2.91$ \\
CSRbDML  & $96.91\pm1.81$ & $96.67\pm3.46$ & $96.92\pm4.15$ \\
JMLC \cite{gao2022joint} & $96.41\pm3.24$ & $98.46\pm1.32$ & $100.0\pm0.00$ \\
NAN \cite{yang2017neural} & $94.62\pm2.70$ & $99.46\pm0.93$ & $100.0\pm0.00$ \\
\hline
nonlocal+DFA \cite{liu2019permutation}& $94.87\pm2.29$& $99.49\pm1.03$ & $100.0\pm0.00$ \\
nonlocal+GAP (baseline) & $78.94\pm1.07$& $81.51\pm1.27$ & $82.69\pm1.18$ \\
\hline
DCSCR(ours)& $\mathbf{97.95\pm1.92}$ & $\mathbf{100.0\pm0.00}$ & $\mathbf{100.0\pm0.00}$ \\
\hline
\end{tabular}}
\end{table}

\begin{table}[t]
\centering
\scriptsize{  % \small, \footnotesize, \scriptsize, \tiny
\caption{Average classification accuracy and standard deviations on MoBo dataset (\%).} \label{MoBo}
\begin{tabular}{|c|c|c|c|}
\hline
\textbf{Methods} & \textbf{50 frames}& \textbf{100 frames}& \textbf{200 frames} \\
\hline
DCC \cite{kim2008canonical} & $78.57\pm4.81$ & $78.10\pm2.24$ & $82.86\pm4.24$ \\
MMD \cite{wang2012manifold} & $79.21\pm7.14$ & $88.25\pm2.83$ & $93.81\pm2.00$ \\
AHISD \cite{cevikalp2010face} & $79.37\pm5.92$ & $81.32\pm5.01$ & $89.22\pm3.61$ \\
CHISD \cite{cevikalp2010face} & $76.42\pm5.74$ & $83.21\pm5.74$  & $90.21\pm6.48$ \\
SANP \cite{hu2012face} & $77.33\pm1.73$ & $87.11\pm4.24$ & $92.13\pm5.83$ \\
RNP \cite{yang2013face} & $85.08\pm3.53$ & $90.95\pm2.70$ & $96.08\pm2.04$ \\
ISCRC \cite{zhu2014image} & $83.97\pm4.24$ & $88.73\pm1.41$ & $96.83\pm2.83$ \\
DLRC \cite{chen2014dual} & $84.44\pm4.79$ & $90.79\pm5.29$ & $85.24\pm4.24$ \\
LEML \cite{huang2015log} & $70.00\pm3.02$ & $72.54\pm4.36$ & $81.32\pm5.66$ \\
PML \cite{huang2015projection} & $73.49\pm3.74$ & $76.67\pm3.32$ & $80.95\pm6.10$ \\
PLRC \cite{feng2016pairwise} & $84.44\pm4.81$ & $90.48\pm2.45$ & $69.84\pm5.00$ \\
DRA \cite{ren2019discriminative} & $84.76\pm3.90$ & $90.16\pm2.68$ & $91.58\pm2.91$ \\
FaceNet \cite{schroff2015facenet} & $89.44\pm4.81$ & $95.36\pm3.69$ & $98.13\pm0.87$ \\
GCR \cite{liu2019group} & $83.73\pm3.21$ & $85.13\pm6.71$ & $93.02\pm4.79$ \\
CSRbDML  & $93.65\pm1.50$ & $96.03\pm2.01$ & $96.83\pm1.83$ \\
JMLC \cite{gao2022joint} & $86.98\pm5.54$ & $94.29\pm0.82$ & $98.41\pm0.01$ \\
NAN \cite{yang2017neural} & $89.60\pm4.79$ & $96.95\pm3.54$ & $98.43\pm1.31$ \\
\hline
nonlocal+DFA \cite{liu2019permutation}& $96.98\pm2.06$& $98.25\pm1.11$ & $98.41\pm0.01$ \\
nonlocal+GAP (baseline) & $79.66\pm0.92$& $82.02\pm1.59$ & $83.77\pm1.53$ \\
\hline
DCSCR(ours)& $\mathbf{98.41\pm1.59}$& $\mathbf{98.57\pm1.32}$ & $\mathbf{98.73\pm0.95}$ \\
\hline
\end{tabular}}
\end{table}

\begin{table}[t]
\centering
\scriptsize{  % \small, \footnotesize, \scriptsize, \tiny
\caption{Comparisons of the average verification accuracy (\%) of our method with baselines and other state-of-the-arts on the YTF dataset.} 
\label{YTF}
\begin{tabular}{|c|c|}
\hline
\textbf{Methods}& \textbf{Accuracy (\%)} \\
\hline
DeepID2+ \cite{sun2015deeply}& $93.20\pm0.20$\\
L-Softmax \cite{liu2016large} & $94.14$\\
Center loss  \cite{wen2016discriminative}& $94.90$\\
Range loss  \cite{zhang2016range}& $93.70$\\
L-GM \cite{wan2018rethinking}& $94.12$\\
CWD loss \cite{zhang2019learning}& $93.76$\\
DAN \cite{rao2019learning}& $94.28\pm0.69$\\
DLMC \cite{wu2020angular}& $94.16$\\
BDML \cite{wei2022discrete}& $93.37$\\
CMCM \cite{sogi2022constrained}& $93.17\pm0.41$\\
\hline
nonlocal+DFA \cite{liu2019permutation}& $92.90\pm1.40$\\
nonlocal+GAP (baseline) & $94.24\pm0.69$\\
\hline
DCSCR(ours)& $\mathbf{94.42\pm0.91}$\\
\hline
\end{tabular}}
\end{table}

\subsection{Evaluation on CMU MoBo Dataset}

On the CMU Mobo dataset, we similarly select first 50, 100, and 200 frames from each set for our experiments, and report the average classification results for ten random experiments in Table \ref{MoBo}. 

As shown in Table \ref{MoBo}, it can be observed that the proposed DCSCR outperforms the comparison methods and the two baseline methods in all cases. Similarly, we find that DLRC and PLRC are sensitive to the size of the image sets, while the recognition accuracy of our method increases steadily with the increase of the number of selected frames. All above observations demonstrate that our proposed DCSCR is more suitable for image set classification. Specifically, on the 50 frames case, our method achieves the best classification performance ($98.41\%$), which reduces the error of sub-optimal method by about $47.19\%$; on the 100 frames case and the 200 frames case, our method also achieves the best classification performance ($98.57\%$ and $98.73\%$, respectively), which reduces the error of sub-optimal method by about $18.29\%$ and $20.13\%$, respectively. In addition, compared with nonlocal+DFA, our method achieves much better performance. Compared with baseline method nonlocal+GAP, our method improves the accuracy by 18.75 percent (50 frames), 16.55 percent (100 frames) and 14.96 percent (200 frames), respectively. All these results demonstrate that combining traditional CSCR method with deep models can indeed improve classification performance.

\begin{figure}[!t]
\centering
\includegraphics[width=0.5\columnwidth]{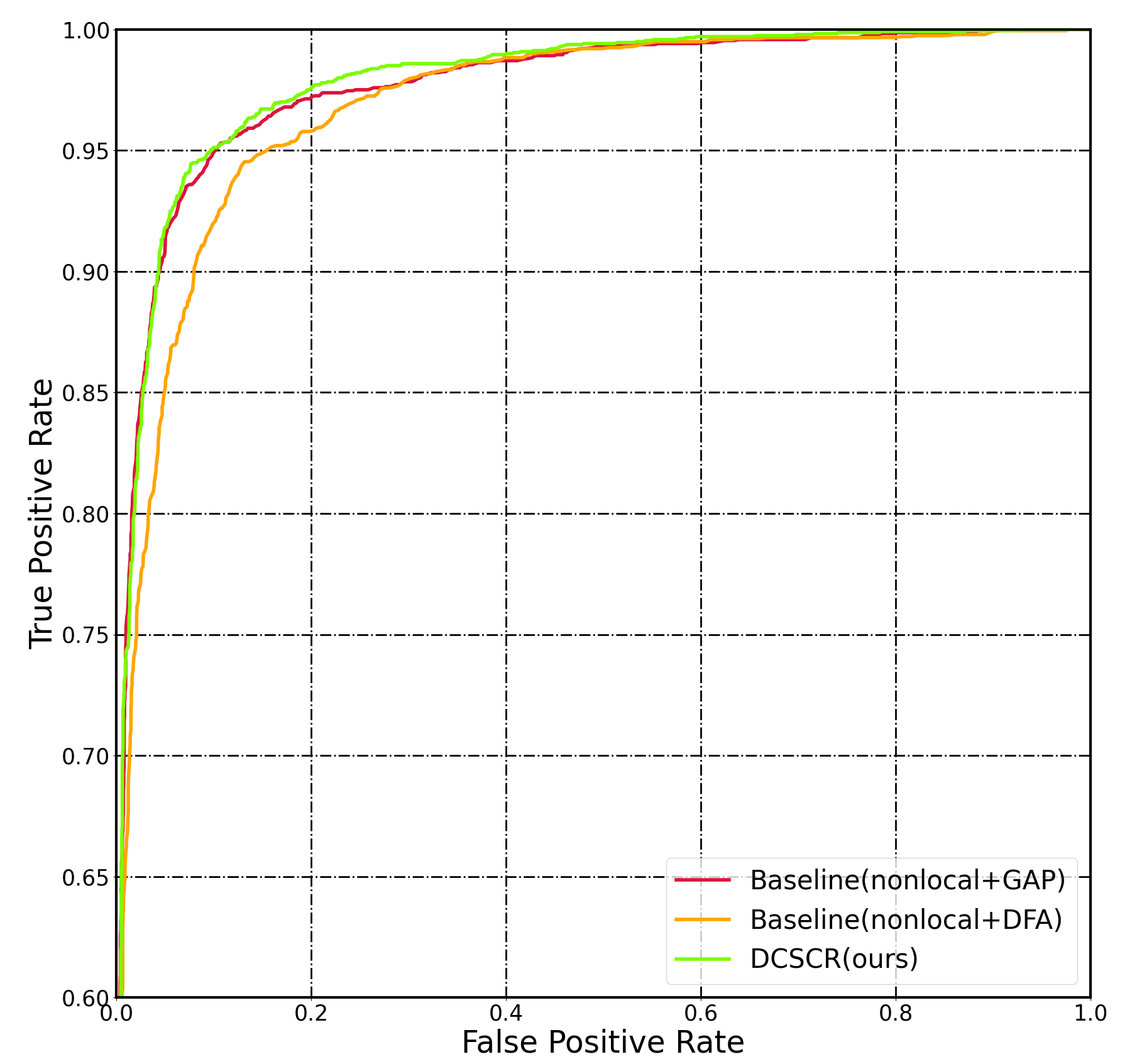}
\caption{Average ROC curves of baseline methods and our DCSCR on the YTF dataset over the 10 folds.}
\label{fig2}
\end{figure}

\subsection{Evaluation on YouTube Face (YTF) dataset}

The video face verification task is aimed at verifying whether two video face clips belong to the same person, which can be considered as a special case of image set classification. In this subsection, we test our method on the YouTube Face (YTF) dataset \cite{wolf2011face}. The average experimental results of our DCSCR, its baseline methods and other methods are shown in Table \ref{YTF}.

From this table, we can see that our DCSCR approach achieves best verification result. Besides, compared our DCSCR (94.42\%) with the baseline nonlocal+GAP (94.24\%), DCSCR achieves a better verification accuracy, which demonstrates that the proposed CSCRMLM is effective; Compared to another baseline nonlocal+DFA (PIFR), DCSCR achieves much better performance, which means that class-specific collaborative representation is better than nonparametric dictionary learning \textcolor{red}{in calculating the concept-level similarity.} In addition, the average ROC curves of our DCSCR and baseline methods over the 10 folds are shown in Fig \ref{fig2}, and the area under the curve of our DCSCR is larger than that of baseline methods in the same setting, which convincingly demonstrates the effectiveness of our methods.

\subsection{Ablation Studies}
To validate the effectiveness of each component, the ablation studies are performed on Honda (50 frames) and MoBo (50 frames) datasets, and the average experimental results are shown in Table \ref{ablation}. Notably, ``Full'' means all components are used, which equals to the proposed DCSCR method.

\begin{table}[!t]
\centering
%\scriptsize  % \small, \footnotesize, \scriptsize, \tiny
\caption{The effects of different components on Honda and MoBo datasets.} 
\label{ablation}
    \begin{tabular}{c c c c  c c }
    \hline
Methods    & DFE & GFLM & CSCRMLM & Honda Acc (\%) & MoBo Acc (\%)  \\
    \hline 
1           & \checkmark &            &            & $75.35\pm2.02$ & $75.78\pm2.45$  \\
2           & \checkmark & \checkmark &            & $78.94\pm1.07$ & $79.66\pm0.92$  \\ 
3           & \checkmark &            & \checkmark & $94.32\pm2.38$ & $95.57\pm1.75$ \\ 
\hline
4(Full)     & \checkmark & \checkmark & \checkmark & $\mathbf{97.95\pm1.92}$ & $\mathbf{98.41\pm1.59}$ \\
\hline
\end{tabular}
\end{table}

From the table, it can be seen that all components have their corresponding roles. When they are used together, we achieve the best classification results. Specifically, compared with ``method 1'' (i.e., only DFE is used), DCSCR improves the accuracy by 22.6 percent and 22.63 percent, respectively, which fully demonstrates the effectiveness of GFLM and CSCRMLM. Compared with ``method 2'', ``method 3'' improves the accuracy by 15.53 percent on Honda dataset, and by 15.91 percent on MoBo dataset, which means that compared to GFLM, CSCRMLM is more important on few-shot classification task. This finding is consistent with our theoretical analysis that convolutional neural networks using cross entropy loss are difficult to handle few-shot image set classification problems, while the CSCRMLM proposed in this paper can effectively address this task. We guess that there are two primary reasons for this phenomenon: First, DFE+GFLM uses the average value of all images in the set as the concept-level features, which makes it struggles to capture the most critical discriminative information. Second, the concept-level features it outputs are fixed and cannot be adaptively adjusted according to the specific requirements of different tasks.

\subsection{Parameter Sensitivity Analysis}

\begin{figure}[!t]
\centering
\subfloat[]{\includegraphics[width=2in]{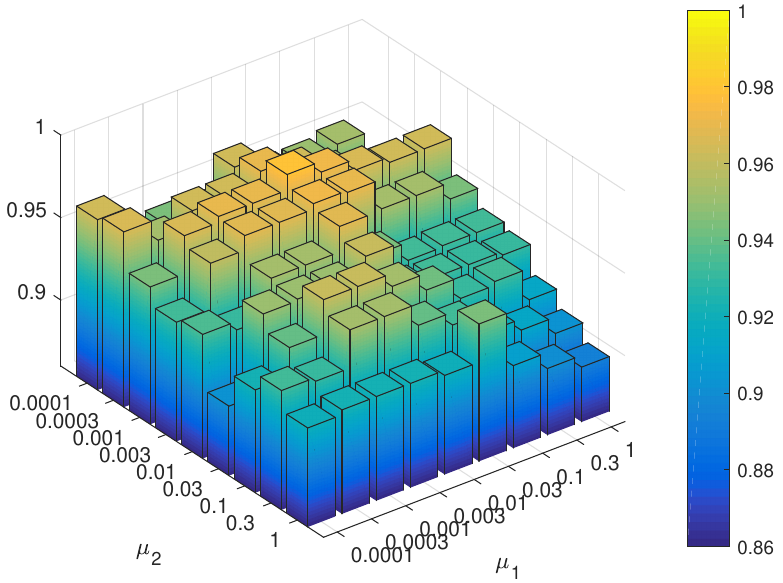}% [width=1.5in,height=0.81in]
\label{fig:tau12}}
\hfil
\subfloat[]{\includegraphics[width=2in]{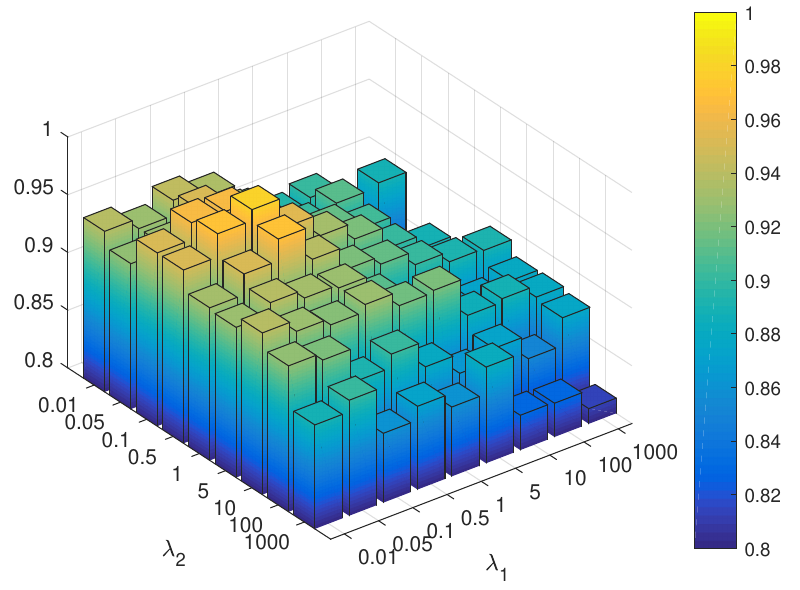}% [width=1.5in,height=0.81in]
\label{fig:t1l2}}
\caption{Effects of different parameters (a) ($\mu_1$, $\mu_2$) and (b) ($\lambda_1$, $\lambda_2$) on Honda (50 frames) dataset.}
\label{fig:diff_parameters}
\end{figure}

In the proposed method, there are four hyper-parameters that may influence its performance: $\mu_1$, $\mu_2$, $\lambda_1$ and $\lambda_2$. So, in this subsection we will verify their influence on ISC accuracy on Honda (50 frames) dataset. Specifically, we first fixe parameters $\lambda_1$, $\lambda_2$, and vary $\mu_1$, $\mu_2$ in the range of [0.001, 1]. The experimental results are displayed in Fig. \ref{fig:diff_parameters}(a). As can be seen from this figure that our DCSCR achieves the best result when ($\mu_1, \mu_2$) = (0.01, 0.001), and it achieves better results when $0.0001 \leq \mu_1 \leq 0.3$, $0.0003 \leq \mu_2 \leq 0.03$. \textcolor{red}{The intrinsic reason for this trend is that: $\mu_1$ controls the intra-class pull force of the contrastive loss, which drives the concept-level features of the same-category image sets to be close to each other. An overly small $\mu_1$ will lead to insufficient intra-class aggregation, while an overly large $\mu_1$ will cause over-fitting to the training set and reduce the generalization ability of the model. For $\mu_2$, it controls the inter-class push force that separates the concept-level features of different-category image sets. A moderate $\mu_2$ ($0.0003\!\sim\!0.03$) can ensure effective inter-class discrimination, while an excessively large $\mu_2$ will over-emphasize the inter-class distance and destroy the intrinsic feature distribution of the image sets, resulting in a sharp drop in classification accuracy. This also explains why the model is slightly sensitive to $\mu_2$---the inter-class push force requires a more precise range to balance discrimination and feature distribution.}

Next, we fix parameters $\mu_1$, $\mu_2$, and vary $\lambda_1$, $\lambda_2$ in the range of [0.01, 1000]. The experimental results are shown in Fig. \ref{fig:diff_parameters}(b). From this figure, we observe that DCSCR achieves the best result when ($\lambda_1, \lambda_2$) = (0.1, 0.5), and it achieves better results when $0.01 \leq \lambda_1 \leq 0.5$, $0.01 \leq \lambda_2 \leq 1$, which means that when the parameters $\lambda_1$, $\lambda_2$ are greater than 1, the proposed DCSCR is somewhat sensitive to them. \textcolor{red}{ The theoretical rationale for this performance trend is: $\lambda_1$ and $\lambda_2$ serve as $l_2$ regularization terms for the collaborative representation weights $\alpha_i$ and $\beta_j$, whose core function is to encourage the model to reconstruct the affine subspace of image sets with fewer high quality and discriminative samples, while suppressing the interference of low quality samples. A moderate regularization coefficient (with $\lambda_1$ in the range of $0.01\!\sim\!0.5$ and $\lambda_2$ in $0.01\!\sim\!1$) can achieve the aforementioned regularization objective and strike a balance between sample selection and feature utilization. When $\lambda_1$ and $\lambda_2$ are greater than 1, the regularization constraint becomes excessively strong, which forces the collaborative representation weights to approach 0. This prevents the model from fully exploring the discriminative information in the image sets and ultimately leads to a significant decline in classification performance, which is also the key reason why the model is sensitive to overly large values of $\lambda_1$ and $\lambda_2$.}

\textcolor{red}{Based on the above experimental results of sensitivity analysis and theoretical mechanism analysis, the optimal values and recommended applicable ranges of each hyper-parameter in this paper are determined as follows: the optimal values are set to $\mu_1\!=\!0.01$, $\mu_2\!=\!0.001$, $\lambda_1\!=\!0.1$ and $\lambda_2\!=\!0.5$. To ensure the discriminability, generalization and robustness of the model in few-shot image set classification tasks, the recommended parameter ranges for stable performance are $\mu \in [0.0003, 0.03]$ (for $\mu_1$ and $\mu_2$) and $\lambda \in [0.01, 0.3]$ (for $\lambda_1$ and $\lambda_2$). }

\begin{table}[!t]
\centering
%\scriptsize  % \small, \footnotesize, \scriptsize, \tiny
\caption{\textcolor{red}{Parameter and computational cost analysis.} }
\label{tab:time}
\textcolor{red}{
    \begin{tabular}{c c c |c  c c }
    \hline
\multicolumn{3}{c|}{Params(M)} & \multicolumn{3}{c}{Flops(G)}  \\
    \hline 
DFE(offline) & GFLM & CSCRMLM & DFE(offline) & GFLM & CSCRMLM    \\
    \hline 
23.55  & 2.09 & 0 & 0 & 0.82$\times$100 & 0.85 \\
\hline
\end{tabular} }
\end{table}

\subsection{\textcolor{red}{Computational Complexity Analysis}}
\textcolor{red}{In this section, we analyze the parameter size and computational cost of each module in the proposed DCSCR method based on Table 5. All results are calculated on the Honda/UCSD dataset with the image set length set to $m=n=50$. Notably, to improve computational efficiency, this study adopts an offline DFE feature extraction strategy: we first compute and pre-store the deep features of all images using the pre-trained DFE module, and only update the GFLM module in the subsequent optimization phase, thus avoiding redundant computations caused by repeated feature extraction. As shown in Table 5, the total parameter size of DCSCR is 25.64M, where the offline DFE contributes 23.55M, the GFLM contributes 2.09M, and the CSCRMLM — a parameter-free metric learning module based on collaborative representation and ADMM alternating optimization — has 0 parameters. This design effectively controls the model complexity while maintaining classification performance. }

\textcolor{red}{In terms of computational cost (FLOPs), the offline DFE incurs 0G FLOPs, the GFLM accounts for $0.82 \times 100 = 82$G FLOPs (the multiplication by 100 is due to the input containing two image sets with a total of 100 images), which is the main computational bottleneck of the model. The CSCRMLM only contributes 0.85G FLOPs, mainly from the linear solution of collaborative representation and inter-class distance calculation, with extremely low computational overhead. Overall, the total computational cost of DCSCR is 82.85G, where the GFLM accounts for approximately 98.97\% and the CSCRMLM only accounts for about 1.03\%. It is worth noting that the GFLM is sensitive to the number of samples in the image set, and its computational cost fluctuates with the change in the number of input images.}

%%%%%%%%%%%%%%%%%%%%%%%%%%%%%%%%%%%%%%%%%%%%%%%%%%%%%%%%%%%%%%%%%%%%%%%%%%%%%%%%

\section{Conclusion} \label{sec:conclusion}
In this paper, we have presented a novel approach called Deep Class-Specific Collaborative Representation for image set classification. DCSCR combines deep networks with class-specific collaborative representation image set classification methods, which can integrate the concept-level modeling advantages of traditional methods with the powerful frame-level feature learning capabilities of deep networks. With a bi-level training process, the method can use a unified way to train the deep learning and collaborative representation learning jointly. \textcolor{red}{Extensive experiments on benchmark datasets show that our DCSCR method achieves SOTA performance in few-shot ISC, with classification accuracies of 97.95\%/100\%/100\% (50/100/200 frames) on Honda/UCSD, 98.41\%/98.57\%/98.73\% (50/100/200 frames) on CMU MoBo. Notably, DCSCR significantly outperforms representative traditional and deep ISC methods, with superior generalization in few-shot scenarios, fully verifying the effectiveness of the CSCRMLM module’s adaptive concept-level modeling mechanism and the rationality of the bi-level training strategy.}

\textcolor{red}{Although the proposed DCSCR method has achieved excellent performance on various benchmark datasets, it still has several limitations worthy of discussion. First, the model performance is highly dependent on the backbone network, and the selection of different backbone networks leads to significant performance variations. Second, the current model still incurs high computational overhead, and its inference efficiency is susceptible to the sample size of image sets. Based on this, two core future research directions are proposed in this paper: first, fusing the proposed CSCRMLM module with more state-of-the-art deep neural networks such as Vision Transformers to further enhance the feature learning capability of DCSCR for image sets in complex unconstrained scenarios; second, designing a lightweight version of the DCSCR model by optimizing the network structure and streamlining redundant computation steps, which can greatly reduce the computational and memory overhead of the method and improve the feasibility of its practical deployment.}

\section*{Declarations}

\begin{itemize}
\item Funding: Not applicable
\item Conflict of interest: The authors declare that they have no conflict of interest. This article does not contain any studies with human participants or animals performed by any of the authors.
\item Ethics approval and consent to participate: Not applicable
\item Consent for publication: All authors have reviewed and approved the final version of the manuscript, and we agree to the terms of publication as outlined by the journal
\item Data availability: The data will be available on reasonable request. 
\item Materials availability: Not applicable
\item Code availability: Not applicable 
\item Author contribution: Xizhan Gao: Method, Writing, Founding; Wei Hu: Review, Revised.
\end{itemize}


%\bibliographystyle{sn-mathphys-num} 
\bibliography{mybibliography}% common bib file

@article{kim2008canonical,
  title={Canonical correlation analysis of video volume tensors for action categorization and detection},
  author={Kim, Tae-Kyun and Cipolla, Roberto},
  journal={IEEE Transactions on Pattern Analysis and Machine Intelligence},
  volume={31},
  number={8},
  pages={1415--1428},
  year={2008},
  publisher={IEEE}
}

@article{wang2012manifold,
  title={Manifold--manifold distance and its application to face recognition with image sets},
  author={Wang, Ruiping and Shan, Shiguang and Chen, Xilin and Dai, Qionghai and Gao, Wen},
  journal={IEEE Transactions on Image Processing},
  volume={21},
  number={10},
  pages={4466--4479},
  year={2012},
  publisher={IEEE}
}

@inproceedings{cevikalp2010face,
  title={Face recognition based on image sets},
  author={Cevikalp, Hakan and Triggs, Bill},
  booktitle={2010 IEEE Computer Society Conference on Computer Vision and Pattern Recognition},
  pages={2567--2573},
  year={2010},
  organization={IEEE}
}

@article{hu2012face,
  title={Face recognition using sparse approximated nearest points between image sets},
  author={Hu, Yiqun and Mian, Ajmal S and Owens, Robyn},
  journal={IEEE transactions on pattern analysis and machine intelligence},
  volume={34},
  number={10},
  pages={1992--2004},
  year={2012},
  publisher={IEEE}
}

@article{zhu2014image,
  title={Image set-based collaborative representation for face recognition},
  author={Zhu, Pengfei and Zuo, Wangmeng and Zhang, Lei and Shiu, Simon Chi-Keung and Zhang, David},
  journal={IEEE transactions on information forensics and security},
  volume={9},
  number={7},
  pages={1120--1132},
  year={2014},
  publisher={IEEE}
}

@inproceedings{chen2014dual,
  title={Dual linear regression based classification for face cluster recognition},
  author={Chen, Liang},
  booktitle={Proceedings of the IEEE conference on computer vision and pattern recognition},
  pages={2673--2680},
  year={2014}
}

@inproceedings{feng2016pairwise,
  title={Pairwise linear regression classification for image set retrieval},
  author={Feng, Qingxiang and Zhou, Yicong and Lan, Rushi},
  booktitle={Proceedings of the IEEE conference on computer vision and pattern recognition},
  pages={4865--4872},
  year={2016}
}

@inproceedings{zhong2019ghostvlad,
  title={Ghostvlad for set-based face recognition},
  author={Zhong, Yujie and Arandjelovi{\'c}, Relja and Zisserman, Andrew},
  booktitle={Computer Vision--ACCV 2018: 14th Asian Conference on Computer Vision, Perth, Australia, December 2--6, 2018, Revised Selected Papers, Part II 14},
  pages={35--50},
  year={2019},
  organization={Springer}
}

@inproceedings{yang2017neural,
  title={Neural aggregation network for video face recognition},
  author={Yang, Jiaolong and Ren, Peiran and Zhang, Dongqing and Chen, Dong and Wen, Fang and Li, Hongdong and Hua, Gang},
  booktitle={Proceedings of the IEEE conference on computer vision and pattern recognition},
  pages={4362--4371},
  year={2017}
}

@article{sogi2022constrained,
  title={Constrained mutual convex cone method for image set based recognition},
  author={Sogi, Naoya and Zhu, Rui and Xue, Jing-Hao and Fukui, Kazuhiro},
  journal={Pattern Recognition},
  volume={121},
  pages={108190},
  year={2022},
  publisher={Elsevier}
}

@inproceedings{hamm2008grassmann,
  title={Grassmann discriminant analysis: a unifying view on subspace-based learning},
  author={Hamm, Jihun and Lee, Daniel D},
  booktitle={Proceedings of the 25th international conference on Machine learning},
  pages={376--383},
  year={2008}
}

@inproceedings{harandi2011graph,
  title={Graph embedding discriminant analysis on Grassmannian manifolds for improved image set matching},
  author={Harandi, Mehrtash T and Sanderson, Conrad and Shirazi, Sareh and Lovell, Brian C},
  booktitle={CVPR 2011},
  pages={2705--2712},
  year={2011},
  organization={IEEE}
}

@inproceedings{huang2015log,
  title={Log-euclidean metric learning on symmetric positive definite manifold with application to image set classification},
  author={Huang, Zhiwu and Wang, Ruiping and Shan, Shiguang and Li, Xianqiu and Chen, Xilin},
  booktitle={International conference on machine learning},
  pages={720--729},
  year={2015},
  organization={PMLR}
}

@article{harandi2017dimensionality,
  title={Dimensionality reduction on SPD manifolds: The emergence of geometry-aware methods},
  author={Harandi, Mehrtash and Salzmann, Mathieu and Hartley, Richard},
  journal={IEEE transactions on pattern analysis and machine intelligence},
  volume={40},
  number={1},
  pages={48--62},
  year={2017},
  publisher={IEEE}
}

@inproceedings{huang2015projection,
  title={Projection metric learning on Grassmann manifold with application to video based face recognition},
  author={Huang, Zhiwu and Wang, Ruiping and Shan, Shiguang and Chen, Xilin},
  booktitle={Proceedings of the IEEE conference on computer vision and pattern recognition},
  pages={140--149},
  year={2015}
}

@inproceedings{chao2019gaitset,
  title={Gaitset: Regarding gait as a set for cross-view gait recognition},
  author={Chao, Hanqing and He, Yiwei and Zhang, Junping and Feng, Jianfeng},
  booktitle={Proceedings of the AAAI conference on artificial intelligence},
  volume={33},
  number={01},
  pages={8126--8133},
  year={2019}
}

@inproceedings{liu2017quality,
  title={Quality aware network for set to set recognition},
  author={Liu, Yu and Yan, Junjie and Ouyang, Wanli},
  booktitle={Proceedings of the IEEE conference on computer vision and pattern recognition},
  pages={5790--5799},
  year={2017}
}

@inproceedings{he2016deep,
  title={Deep residual learning for image recognition},
  author={He, Kaiming and Zhang, Xiangyu and Ren, Shaoqing and Sun, Jian},
  booktitle={Proceedings of the IEEE conference on computer vision and pattern recognition},
  pages={770--778},
  year={2016}
}

@inproceedings{szegedy2016rethinking,
  title={Rethinking the inception architecture for computer vision},
  author={Szegedy, Christian and Vanhoucke, Vincent and Ioffe, Sergey and Shlens, Jon and Wojna, Zbigniew},
  booktitle={Proceedings of the IEEE conference on computer vision and pattern recognition},
  pages={2818--2826},
  year={2016}
}

@inproceedings{wang2018non,
  title={Non-local neural networks},
  author={Wang, Xiaolong and Girshick, Ross and Gupta, Abhinav and He, Kaiming},
  booktitle={Proceedings of the IEEE conference on computer vision and pattern recognition},
  pages={7794--7803},
  year={2018}
}

@inproceedings{liu2019permutation,
  title={Permutation-invariant feature restructuring for correlation-aware image set-based recognition},
  author={Liu, Xiaofeng and Guo, Zhenhua and Li, Site and Kong, Lingsheng and Jia, Ping and You, Jane and Kumar, BVK},
  booktitle={Proceedings of the IEEE/CVF International Conference on Computer Vision},
  pages={4986--4996},
  year={2019}
}

@article{gao2022joint,
  title={Joint metric learning-based class-specific representation for image set classification},
  author={Gao, Xizhan and Niu, Sijie and Wei, Dong and Liu, Xingrui and Wang, Tingwei and Zhu, Fa and Dong, Jiwen and Sun, Quansen},
  journal={IEEE Transactions on Neural Networks and Learning Systems},
  volume={35},
  number={5},
  pages = {6731-6745},
  year={2024},
  publisher={IEEE}
}

@inproceedings{yang2013face,
  title={Face recognition based on regularized nearest points between image sets},
  author={Yang, Meng and Zhu, Pengfei and Van Gool, Luc and Zhang, Lei},
  booktitle={2013 10th IEEE international conference and workshops on automatic face and gesture recognition (FG)},
  pages={1--7},
  year={2013},
  organization={IEEE}
}

@inproceedings{schroff2015facenet,
  title={Facenet: A unified embedding for face recognition and clustering},
  author={Schroff, Florian and Kalenichenko, Dmitry and Philbin, James},
  booktitle={Proceedings of the IEEE conference on computer vision and pattern recognition},
  pages={815--823},
  year={2015}
}

@article{ren2019discriminative,
  title={Discriminative residual analysis for image set classification with posture and age variations},
  author={Ren, Chuan-Xian and Luo, You-Wei and Xu, Xiao-Lin and Dai, Dao-Qing and Yan, Hong},
  journal={IEEE Transactions on Image Processing},
  volume={29},
  pages={2875--2888},
  year={2019},
  publisher={IEEE}
}

@article{liu2019group,
  title={Group collaborative representation for image set classification},
  author={Liu, Bo and Jing, Liping and Li, Jia and Yu, Jian and Gittens, Alex and Mahoney, Michael W},
  journal={International Journal of Computer Vision},
  volume={127},
  pages={181--206},
  year={2019},
  publisher={Springer}
}

@inproceedings{lee2003video,
  title={Video-based face recognition using probabilistic appearance manifolds},
  author={Lee, Kuang-Chih and Ho, Jeffrey and Yang, Ming-Hsuan and Kriegman, David},
  booktitle={2003 IEEE Computer Society Conference on Computer Vision and Pattern Recognition, 2003. Proceedings.},
  volume={1},
  pages={I--I},
  year={2003},
  organization={IEEE}
}

@book{gross2001cmu,
  title={The cmu motion of body (mobo) database},
  author={Gross, Ralph},
  year={2001},
  publisher={Carnegie Mellon University, The Robotics Institute}
}

@inproceedings{wolf2011face,
  title={Face recognition in unconstrained videos with matched background similarity},
  author={Wolf, Lior and Hassner, Tal and Maoz, Itay},
  booktitle={CVPR 2011},
  pages={529--534},
  year={2011},
  organization={IEEE}
}

@inproceedings{wu2020angular,
  title={Angular discriminative deep feature learning for face verification},
  author={Wu, Bowen and Wu, Huaming},
  booktitle={ICASSP 2020-2020 IEEE International Conference on Acoustics, Speech and Signal Processing (ICASSP)},
  pages={2133--2137},
  year={2020},
  organization={IEEE}
}

@inproceedings{wan2018rethinking,
  title={Rethinking feature distribution for loss functions in image classification},
  author={Wan, Weitao and Zhong, Yuanyi and Li, Tianpeng and Chen, Jiansheng},
  booktitle={Proceedings of the IEEE conference on computer vision and pattern recognition},
  pages={9117--9126},
  year={2018}
}

@article{zhang2019learning,
  title={Learning deep discriminative face features by customized weighted constraint},
  author={Zhang, Monica MY and Shang, Kun and Wu, Huaming},
  journal={Neurocomputing},
  volume={332},
  pages={71--79},
  year={2019},
  publisher={Elsevier}
}

@inproceedings{sun2015deeply,
  title={Deeply learned face representations are sparse, selective, and robust},
  author={Sun, Yi and Wang, Xiaogang and Tang, Xiaoou},
  booktitle={Proceedings of the IEEE conference on computer vision and pattern recognition},
  pages={2892--2900},
  year={2015}
}

@article{rao2019learning,
  title={Learning discriminative aggregation network for video-based face recognition and person re-identification},
  author={Rao, Yongming and Lu, Jiwen and Zhou, Jie},
  journal={International Journal of Computer Vision},
  volume={127},
  pages={701--718},
  year={2019},
  publisher={Springer}
}

@article{wei2022discrete,
  title={Discrete metric learning for fast image set classification},
  author={Wei, Dong and Shen, Xiaobo and Sun, Quansen and Gao, Xizhan},
  journal={IEEE Transactions on Image Processing},
  volume={31},
  pages={6471--6486},
  year={2022},
  publisher={IEEE}
}

@inproceedings{wen2016discriminative,
  title={A discriminative feature learning approach for deep face recognition},
  author={Wen, Yandong and Zhang, Kaipeng and Li, Zhifeng and Qiao, Yu},
  booktitle={Computer Vision--ECCV 2016: 14th European Conference, Amsterdam, The Netherlands, October 11--14, 2016, Proceedings, Part VII 14},
  pages={499--515},
  year={2016},
  organization={Springer}
}

@article{zhang2016range,
  title={Range loss for deep face recognition with long-tail},
  author={Zhang, Xiao and Fang, Zhiyuan and Wen, Yandong and Li, Zhifeng and Qiao, Yu},
  journal={arXiv preprint arXiv:1611.08976},
  year={2016}
}

@article{liu2016large,
  title={Large-margin softmax loss for convolutional neural networks},
  author={Liu, Weiyang and Wen, Yandong and Yu, Zhiding and Yang, Meng},
  journal={arXiv preprint arXiv:1612.02295},
  year={2016}
}

@article{wen2010alternating,
  title={Alternating direction augmented Lagrangian methods for semidefinite programming},
  author={Wen, Zaiwen and Goldfarb, Donald and Yin, Wotao},
  journal={Mathematical Programming Computation},
  volume={2},
  number={3-4},
  pages={203--230},
  year={2010},
  publisher={Springer}
}

@article{wang2022learning,
  title={Learning a discriminative SPD manifold neural network for image set classification},
  author={Wang, Rui and Wu, Xiao-Jun and Chen, Ziheng and Xu, Tianyang and Kittler, Josef},
  journal={Neural networks},
  volume={151},
  pages={94--110},
  year={2022},
  publisher={Elsevier}
}

@article{wang2023u,
  title={U-SPDNet: An SPD manifold learning-based neural network for visual classification},
  author={Wang, Rui and Wu, Xiao-Jun and Xu, Tianyang and Hu, Cong and Kittler, Josef},
  journal={Neural networks},
  volume={161},
  pages={382--396},
  year={2023},
  publisher={Elsevier}
}

@inproceedings{arandjelovic2005face,
  title={Face recognition with image sets using manifold density divergence},
  author={Arandjelovic, Ognjen and Shakhnarovich, Gregory and Fisher, John and Cipolla, Roberto and Darrell, Trevor},
  booktitle={2005 IEEE Computer Society Conference on Computer Vision and Pattern Recognition (CVPR'05)},
  volume={1},
  pages={581--588},
  year={2005},
  organization={IEEE}
}

@article{yan2019multiple,
  title={Multiple kernel dimensionality reduction based on linear regression virtual reconstruction for image set classification},
  author={Yan, Wenzhu and Sun, Quansen and Sun, Huaijiang and Li, Yanmeng and Ren, Zhenwen},
  journal={Neurocomputing},
  volume={361},
  pages={256--269},
  year={2019},
  publisher={Elsevier}
}

@article{Zhao2024VMHI,
  title={Multi-view hyperspectral image classification via weighted sparse representation},
  author={Yue Zhao and Yao Qin and Zhifei Li and Wenxin Huang and Rui Hou },
  journal={Multimedia Tools and Applications},
  volume={83},
  pages={90207–90226},
  year={2024},
  publisher={Springer}
}

@article{Guan2024RGMMC,
  title={Robust Grassmann manifold convex hull collaborative representation learning and its kernel extension for image set analysis},
  author={Yao Guan and Jiayi Yao and Wenzhu Yan and Yanmeng Li},
  journal={Multimedia Systems},
  volume={30},
  pages={322},
  year={2024},
  publisher={Springer}
}

@article{shukla2025fraudulent,
  title={Fraudulent account detection in social media using hybrid deep transformer model and hyperparameter optimization},
  author={Shukla, Prashant Kumar and Veerasamy, Bala Dhandayuthapani and Alduaiji, Noha and Addula, Santosh Reddy and Pandey, Ankur and Shukla, Piyush Kumar},
  journal={Scientific Reports},
  volume={15},
  number={1},
  pages={38447},
  year={2025},
  publisher={Nature Publishing Group UK London}
}

@ARTICLE{Wang2024SMDML,
  author={Wang, Rui and Wu, Xiao-Jun and Chen, Ziheng and Hu, Cong and Kittler, Josef},
  journal={IEEE Transactions on Neural Networks and Learning Systems}, 
  title={SPD Manifold Deep Metric Learning for Image Set Classification}, 
  year={2024},
  volume={35},
  number={7},
  pages={8924-8938}
}

@ARTICLE{Wang2024DMLSM,
  author={Wang, Rui and Wu, Xiao-Jun and Xu, Tianyang and Hu, Cong and Kittler, Josef},
  journal={IEEE Transactions on Circuits and Systems for Video Technology}, 
  title={Deep Metric Learning on the SPD Manifold for Image Set Classification}, 
  year={2024},
  volume={34},
  number={2},
  pages={663-680}
}

@ARTICLE{Wang2022SymNet,
  author={Wang, Rui and Wu, Xiao-Jun and Kittler, Josef},
  journal={IEEE Transactions on Neural Networks and Learning Systems}, 
  title={SymNet: A Simple Symmetric Positive Definite Manifold Deep Learning Method for Image Set Classification}, 
  year={2022},
  volume={33},
  number={5},
  pages={2208-2222}
}

@ARTICLE{Wang2022MRMD,
  author={Wang, Rui and Wu, Xiao-Jun and Chen, Kai-Xuan and Kittler, Josef},
  journal={IEEE Transactions on Big Data}, 
  title={Multiple Riemannian Manifold-Valued Descriptors Based Image Set Classification With Multi-Kernel Metric Learning}, 
  year={2022},
  volume={8},
  number={3},
  pages={753-769}
}

\end{document}